\documentclass[journal]{IEEEtran}
\usepackage{cite}
\usepackage{enumitem}
\usepackage{booktabs}
\usepackage{makecell}
\usepackage{multirow}
\usepackage{soul}
\usepackage{array}
\usepackage{graphicx} 
\usepackage{amsthm,amsmath,amssymb,amsfonts}
\usepackage{dsfont}
\usepackage{algorithm,algorithmic}

\theoremstyle{definition}

\usepackage{textcomp}
\usepackage[dvipsnames]{xcolor}
\usepackage{fancyhdr}
\def\BibTeX{{\rm B\kern-.05em{\sc i\kern-.025em b}\kern-.08em
    T\kern-.1667em\lower.7ex\hbox{E}\kern-.125emX}}

\DeclareRobustCommand*{\IEEEauthorrefmark}[1]{
  \raisebox{0pt}[0pt][0pt]{\textsuperscript{\footnotesize #1}}%
}

\begin{document}
\title{LLM2TEA: An Agentic AI Designer for Discovery with Generative Evolutionary Multitasking}

\author{
\IEEEauthorblockN{Melvin~Wong\IEEEauthorrefmark{1}, 
Jiao~Liu\IEEEauthorrefmark{1},
Thiago~Rios\IEEEauthorrefmark{2},
Stefan~Menzel\IEEEauthorrefmark{2},
Yew~Soon~Ong\IEEEauthorrefmark{1,3}} \\ \, \\
\IEEEauthorblockA{\textit{\IEEEauthorrefmark{1}College of Computing \& Data Science (CCDS), Nanyang Technological University (NTU), Singapore} \\
\textit{\IEEEauthorrefmark{2}Honda Research Institute Europe (HRI-EU), Offenbach am Main, Germany} \\
\textit{\IEEEauthorrefmark{3}Centre for Frontier AI Research (CFAR), Agency for Science, Technology and Research (A*STAR), Singapore}  \\ \, \\
\normalsize \{wong1357, jiao.liu, asysong\}@ntu.edu.sg, \{thiago.rios, stefan.menzel\}@honda-ri.de
}
\thanks{\textcopyright 2025 IEEE. Personal use of this material is permitted.  Permission from IEEE must be obtained for all other uses, in any current or future media, including reprinting/republishing this material for advertising or promotional purposes, creating new collective works, for resale or redistribution to servers or lists, or reuse of any copyrighted component of this work in other works.}
}

\maketitle

\begin{abstract}
This paper presents LLM2TEA, a Large Language Model (LLM) driven MultiTask Evolutionary Algorithm, representing the first agentic AI designer of its kind operating with generative evolutionary multitasking (GEM). LLM2TEA enables the crossbreeding of solutions from multiple domains, fostering novel solutions that transcend disciplinary boundaries. Of particular interest is the ability to discover designs that are both novel and conforming to real-world physical specifications. LLM2TEA comprises an LLM to generate genotype samples from text prompts describing target objects, a text-to-3D generative model to produce corresponding phenotypes, a classifier to interpret its semantic representations, and a computational simulator to assess its physical properties. Novel LLM-based multitask evolutionary operators are introduced to guide the search towards high-performing, practically viable designs. Experimental results in conceptual design optimization validate the effectiveness of LLM2TEA, showing \(97\%\) to \(174\%\) improvements in the diversity of novel designs over the current text-to-3D baseline. Moreover, over \(73\%\) of the generated designs outperform the top \(1\%\) of designs produced by the text-to-3D baseline in terms of physical performance. The designs produced by LLM2TEA are not only aesthetically creative but also functional in real-world contexts. Several of these designs have been successfully 3D printed, demonstrating the ability of our approach to transform AI-generated outputs into tangible, physical designs. These designs underscore the potential of LLM2TEA as a powerful tool for complex design optimization and discovery, capable of producing novel and physically viable designs.
\end{abstract}


\section{Introduction} \label{sec:intro}
Generative Artificial Intelligence (GenAI) has demonstrated remarkable capabilities in producing diverse creative designs from free-form natural language prompts~\cite{wong2023promptevo}. Notable examples include the synthesis of creative texts, images, audio, and videos through tools such as SARD~\cite{radwan2024sard}, Stable Diffusion~\cite{podell2023sdxl}, Make-An-Audio~\cite{huang2023make}, and Veo 2.0~\cite{veo2}. Over the past few years, a surge of research in GenAI has focused on enabling the production of high-quality digital media that closely aligns with user preferences through the use of tailored prompts. Despite these widely publicized advancements in digital media, a notable gap remains in applying GenAI to complex engineering design, with comparatively fewer reports addressing its potential within the science and engineering domains.

In the realm of traditional engineering design, which is the primary focus of this work, the promising potential of GenAI in conceptual aerodynamic design has been demonstrated~\cite{rios2023LLM,wong2024generative}. The major challenge identified in applying GenAI to science and engineering lies not only in generating digital designs exhibiting high novelties, but also in ensuring that the designs can translate into tangible physical solutions that adhere to natural principles~\cite{har2012advances} or satisfy specific problem contexts and design specifications~\cite{regenwetter2022deep}. Our early research studies have attempted to address this challenge by advancing GenAI for engineering design optimization~\cite{martins2021engineering,xu2024towards,xu2024precise}. Preliminary results leveraging conventional optimization strategies indicate encouraging progress, particularly through the use of text prompt search to improve aerodynamic performance~\cite{rios2023LLM} while maintaining semantic conformity~\cite{wong2024generative}. These efforts underscore the potential of GenAI to bridge the gap between digital innovation and practical applicability, particularly in engineering disciplines. However, navigating the very high-dimensional language search space often demands an integration of symbolic, statistical, and experiential knowledge to support broader generalization and knowledge transfer. This imperative highlights the need for agentic AI approaches that can engage in objective-directed reasoning and guidance as well as adaptive problem-solving.

To this end, a large language model (LLM) driven MultiTask Evolutionary Algorithm (LLM2TEA), is introduced as the agentic AI designer implemented with generative evolutionary multitasking (GEM), structured to facilitate the cross-pollination of solutions across multiple domains. The objective of LLM2TEA is the discovery of novel solutions that transcend disciplinary boundaries while adhering to natural principles and meeting user specifications. LLM2TEA is composed of an LLM-guided multitask evolutionary algorithm, a text-to-3D generative model responsible for phenotype production, a classifier for interpreting semantic representations, and a physics simulation model to evaluate the physical properties of the generated designs. By solving multiple tasks simultaneously, where each task corresponds to a single-domain search, the exploration of design space is expanded beyond isolated domains and into their intersections, allowing for the emergence of cross-domain designs enabled by knowledge transfer.

Several novel LLM-based multitask evolutionary operators are introduced, incorporating new instruction sets to guide the evolutionary processes. This includes an initialization operator for generating an initial population of genotypes and multitask mating operators, such as crossover, mutation, and selection, to enable the efficient exchange of traits across domains as evolution progresses with the generation of offspring. The resulting genotype population serves as input to the text-to-3D generative model for synthesizing phenotypes. The survival of the best-performing designs is evaluated based on the practical viability of these phenotypes, considering criteria that include physical properties and semantic meanings. In summary, the contributions of this paper are as follows:

\begin{itemize}[leftmargin=1em]
    \item This paper introduces LLM2TEA, a novel agentic AI designer with a generative evolutionary multitasking (GEM) that promotes the crossbreeding of designs from multiple domains, leading to novel solutions that transcend individual domains.
    
    \item LLM2TEA is an LLM-driven multitask evolutionary algorithm with operators synthesizing creative text prompts as guides to an auxiliary text-to-3D generative model in producing corresponding novel 3D designs. LLM2TEA represents the first method operating with the GEM paradigm to leverage the capabilities of LLMs for complex design optimization and the discovery of novel digital designs. These digital designs not only exhibit high novelty but are also translatable into tangible physical solutions that conform to natural laws and fulfill specific problem contexts and design specifications.

    \item A series of complex design and discovery tasks, modeled after real-world design optimization scenarios, is conducted to assess the efficacy of the proposed LLM2TEA. Experimental results indicate that LLM2TEA consistently outperforms the baseline text-to-3D generative model in uncovering novel designs and/or enhancing aerodynamic performance. The proposed novelty score metric is employed for evaluation, with findings showing that over \(45\%\) of generated designs are identified as novel, exceeding the mean performance of the baseline.

    \item It is demonstrated that several designs discovered through LLM2TEA are physically realizable and successfully 3D printed using off-the-shelf stereolithography (SLA) technology. As previously noted, direct outputs from baseline text-to-3D generative models often result in designs that are highly fragmented for physical fabrication. In contrast, the findings indicate that LLM2TEA is capable of identifying physically feasible designs suitable for SLA printing, highlighting its potential as a viable approach for practical rapid design prototyping.
\end{itemize}

\noindent The remainder of the paper is organized as follows: Section II reviews related work from the literature and provides an overview of key background concepts. Section III introduces LLM2TEA in detail. Section IV presents the experimental setup, and the results and discussions are detailed in Section V. Finally, Section VI concludes this paper, highlighting some potential future directions.

\section{Related Works}
This section presents an overview of related work. First, the applications of \textit{text-to-X} generative models (where \textit{X} represents various outputs such as languages, images, or 3D shapes) are reviewed in the context of creative and engineering conceptual design. Next, prompt engineering and search techniques are examined, with a focus on methods to generate effective prompts for text-to-\textit{X} generative models. Finally, recent advancements in the evolutionary multitasking (EMT) paradigm are discussed, with an emphasis on its potential to address cross-domain optimization tasks simultaneously.

\subsection{Text-to-X Generative Models in Creative \& Engineering Conceptual Design}
Recently, text-to-\textit{X} generative models have been introduced to allow users to specify design specifications using free-form natural language prompts. Pioneering work investigated the use of text-to-image generative models to synthesize novel creative digital artifacts that satisfy user preferences \cite{wong2023promptevo}. On the other hand, Point-E \cite{Nichol2022} and Shap-E \cite{Jun2023} were among the first models to generate 3D artifacts from text prompts. In contrast to Point-E, which is based on 3D point cloud representations, Shap-E includes additional features, such as a differentiable implementation of the marching cubes 33 algorithm \cite{cubes1995construction} to synthesize 3D polygonal meshes, which are commonly used representation format in computer graphics or for computational simulation-based applications.

\subsection{Prompt Engineering and Search}
It has been observed that prompt augmentation significantly influences the feasibility and quality of artifacts generated by text-to-\textit{X} generative models. Consequently, designing effective prompts is crucial for fully harnessing the full potential of these generative models for solving diverse tasks. Several studies have proposed prompting techniques based on handcrafted prompt templates to better align the outputs of the model with target objectives~\cite{wei2022chain,yao2024tree}. In contrast, prompt search methods have considered optimization approaches to finding optimal prompts through iterative refinement~\cite{diao2022black,pryzant2023automatic,wen2024hard}.

In recent years, prompt search techniques have also been applied in engineering design optimization \cite{rios2023LLM}. Rios et al. proposed a fully automated evolutionary design optimization framework incorporating evolutionary strategies with performance-based guidance to seek for novel 3D car designs using the Shap-E text-to-3D generative model. However, searching within the text-prompt space can often lead to ill-defined design outputs, as Shap-E may not be explicitly trained on a dataset focused on car shapes. To address this limitation, vision-language models~\cite{pmlr-v139-radford21a,li2023blip} were introduced into the framework to penalize semantically incoherent or structurally implausible designs~\cite{wong2024generative}. The research demonstrated the feasibility of searching effective text prompts to generate designs that meet both practical constraints and visual requirements with evolutionary approaches. However, the resulting design often retained strong resemblances to conventional car shapes, since the optimization process was narrowly targeted at car design objectives.

Due to recent advancements in language comprehension and generation capabilities, LLMs have become effective tools not only for generating prompts but also for serving as optimizers. For instance, Yang \textit{et al.}~\cite{yang2023large} directly utilized an LLM as the optimizer in the prompt optimization task, demonstrating both the feasibility and effectiveness of such an approach. Guo \textit{et al.}~\cite{guo2023connecting} combined evolutionary algorithms with an LLM to enhance prompt optimization, leveraging the strengths of both methods. However, most existing approaches primarily focus on optimizing prompts for a single task. As discussed in Section~\ref{sec:intro}, it is believed that incorporating cross-domain learning into the design optimization process can identify prompts that lead to novel design solutions. Therefore, prompt search via evolutionary multitasking techniques is proposed. Such an approach leverages knowledge from multiple domains to explore a broader design space that extends beyond individual domains and embraces its overlap, prompting the discovery of novel cross-domain designs through knowledge transfer.

\begin{figure*}
    \centering
    \includegraphics[width=1.00\textwidth]{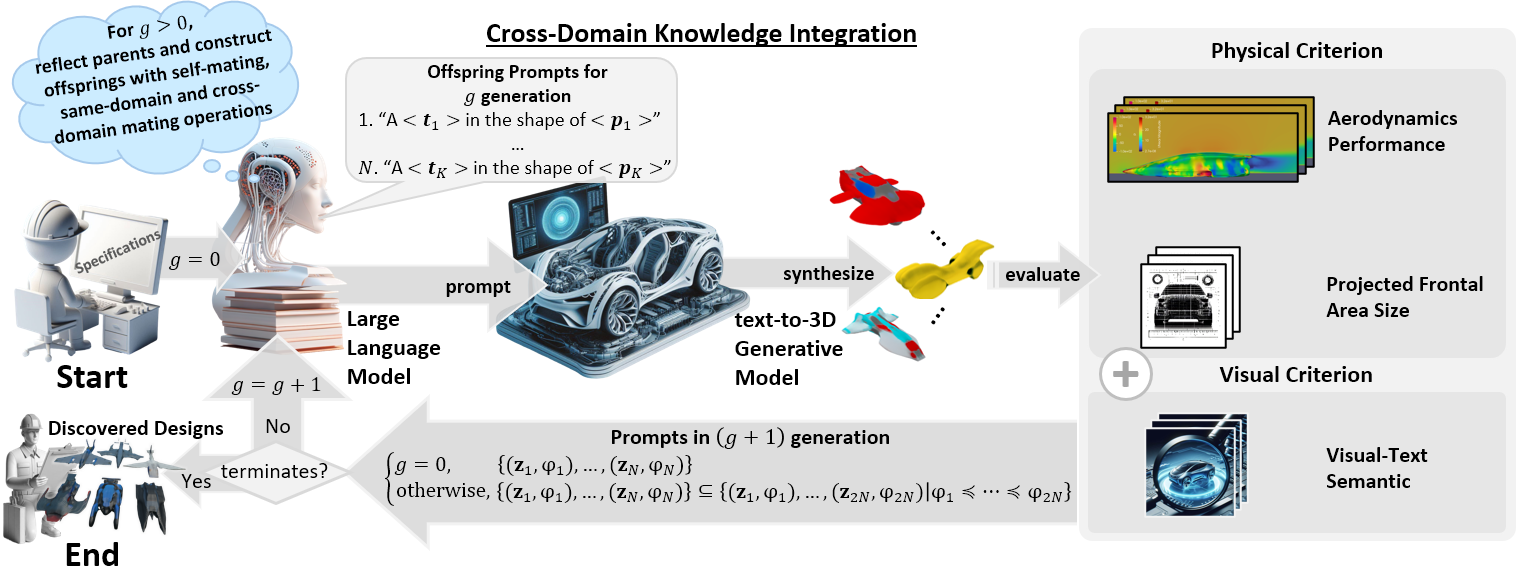}
    \caption{LLM2TEA is a novel \textbf{LLM}-driven \textbf{M}ulti\textbf{T}ask \textbf{E}volutionary \textbf{A}lgorithm for complex design and discovery. LLM2TEA optimizes multiple tasks simultaneously, leveraging cross-domain knowledge learning to drive LLM in evolving prompts. The prompts serve as guides for an auxiliary text-to-3D generative model, expanding exploration into cross-domain regions.}
    \label{fig:framework}
\end{figure*}

\subsection{Evolutionary Multitasking}
Recently, the evolutionary multitasking (EMT) paradigm has emerged as a population-based search methodology designed to handle multiple optimization tasks concurrently~\cite{gupta2017insights}. In contrast to traditional evolutionary algorithms, EMT exploits latent synergies between distinct yet related optimization tasks, thereby enhancing convergence performance. Over the past few years, several successful applications of EMT have been documented across various domains, including manufacturing process design~\cite{liu2024evolutionary}, mobile edge computing~\cite{yang2023evolutionary}, and vehicle routing~\cite{9446541,feng2020explicit}. These examples highlight the potential of EMTs for efficiently solving complex, multifaceted problems by leveraging cross-task knowledge sharing.

While there is surging research in using LLMs as evolutionary optimizers to address both traditional optimization~\cite{liu2023LLMEvoOptimizers,liu2023moead} and prompt optimization tasks~\cite{guo2023connecting}, no documented studies have explored exploiting the contrasting characteristics of multiple domains to enable crossbreeding of solutions with evolutionary multitasking, resulting in the discovery of novel artifacts with hybrid features. This paper aims to investigate this possibility, proposing a novel prompt search technique that operates with generative evolutionary multitasking (GEM). The proposed method not only demonstrates effective optimization but also promotes the discovery of novel solutions.

\section{Agentic AI Designer With Generative Evolutionary Multitasking (GEM)}
In this section, the proposed LLM2TEA is presented. First, the multitask optimization problem of interest is described. Next, each distinctive feature of the LLM2TEA, describing its role in the multitask evolutionary algorithm, is examined in detail.

\subsection{Preliminaries}\label{sec:problem_setting}
Let \(\mathcal{Z} \subseteq \mathbb{Z}^{L \times V}_{\geq 0} \) denote the search space of sequential tokens, where \(L\) is the maximum sequence tokens length and \(V\) is the vocabulary size. Given a set of tasks $\{T_{1},...,T_{K}\}$, the LLM constructs a genotype (prompt) (\(\textbf{z}_{k} \in \mathcal{Z}\)) relevant to the $k$-th task using the following template: 
\\

\centerline{$\textbf{z}_{k} = \text{tokenize}($``A $< \textbf{t}_{k} >$ in the shape of $< \boldsymbol{\rho}_{k} >$''.)}

\noindent \\ Here, a user provides a \textit{target domain} ``$\textbf{t}_{k}$'' that defines the task of interest (e.g., ``$\textbf{t}_{k}$'' as ``car'', ``airplane'' or others). The variable $\boldsymbol{\rho}_{k}$ represents the part of the prompt that LLM2TEA will optimize. Based on this setting, the single-objective function for the $k$-th task is formulated as follows:
\begin{equation} \label{eq:obj}
    \begin{aligned}        
        & \forall T_k, k \in \{1, \ldots, K\}, \\
        & \min_{\boldsymbol{\textbf{z}}_{k} \in \mathcal{Z}}: \mathop{\mathbb{E}}_{p_{\theta}(\textbf{x} | \textbf{z}_{k})} \left[  (1 - \alpha) f_{phy}\left( \textbf{x}\right) + \alpha (1 - f_{visual}(\textbf{x} | \textbf{t}_{k})) \right], \\
    \end{aligned}
\end{equation}
where $p_{\theta}(\textbf{x}|\textbf{z}_i)$ represents a \textit{text-to-3D} model that generates the phenotypes (3D shapes) $\textbf{x}$ based on the prompt $\textbf{z}_{k}$. The function $f_{phy}(\cdot)$ is an evaluator that assesses the shape $\textbf{x}$ against a target physical criterion. Additionally, LLM2TEA visually assesses the shape $\textbf{x}$ with a vision-language model $f_{visual}(\cdot|\textbf{t}_{i})$ by measuring the probability that the generated 3D design visually aligns with the target domain $\textbf{t}_{k}$. The $\alpha$ hyperparameter is introduced as a predefined weight to balance the visual fitness score versus the physical fitness score.

Correspondingly, a multi-objective optimization problem with $M$ objectives can be formulated as follows:
\begin{equation} \label{eq:moo_obj}
    \begin{aligned}       
        & \forall T_k, k \in \{1, \ldots, K\}, \\
        & \min_{\boldsymbol{\textbf{z}}_{k} \in \mathcal{Z}}: \Biggl\{\mathop{\mathbb{E}}_{p_{\theta}(\textbf{x} | \textbf{z}_{k})} [f_{phy}^{(1)}( \textbf{x})],.., \mathop{\mathbb{E}}_{p_{\theta}(\textbf{x} | \textbf{z}_{k})} [f_{phy}^{(M)}( \textbf{x})] \Biggr\} \\
        &\,\,\,\,\, \text{s.t.} \,\,\,\, \underline{c} \leq f_{visual}(\textbf{x} | \textbf{t}_{k}) \leq \bar{c}, \\ 
    \end{aligned}
\end{equation}
where \(\underline{c}\) and \(\bar{c}\) control the exploration of designs within and near the domain boundary, and solutions trade-offs among the multiple criteria are of interest. The setting of these objective functions ensures that the optimized prompt generates shapes that not only meet the target physical criteria but also resemble the conceptual elements of the target domain represented by $\textbf{t}_{k}$.

\subsection{LLM2TEA Overview}
Given the very high-dimensional nature of the language search space, assume the existence of a function \(g: \mathbb{Z}^{(C \times V_{c})^{H}} \rightarrow \mathbb{Z}^{L \times V} \) that can construct the prompt \(\textbf{z}_{k}\), where \(H\) denotes the historical context, \(C\) is the maximum sequence context length and \(V_{c}\) is the context vocabulary size. Further, the hidden Markov property is assumed to hold, which states that all necessary information to derive an outcome exists in the current state (or context in our case), thereby reducing \(H=1\). Under these assumptions, the capabilities of an LLM \(p_{\text{llm}}(\textbf{z}_k | \textbf{t}_k)\) are utilized as the function \(g\) to construct effective prompts that guide the text-to-3D model in generating aerodynamically performing vehicle designs. With these considerations, LLM2TEA (Fig.~\ref{fig:framework}) comprises an LLM that constructs prompts from a set of instructions (depicted in Fig.~\ref{fig:llm}) containing the user specifications embedded in $\textbf{z}_{k}$. The generated prompts are passed to a text-to-3D generative model that synthesizes designs, which are then evaluated against a set of criteria to compute the fitness with respect to the target objective. Details of the key algorithmic features of LLM2TEA for various proposed approaches in complex design and discovery are elaborated upon in this section.

\subsubsection{Population Initialization}\label{sec:initialize} LLM2TEA begins with instructing an LLM to generate an initial population of genotypes, each genotype being a text prompt based on free-form natural language as discussed in Section~\ref{sec:problem_setting}. Specifically, the LLM is presented with a description of the optimization problem (shown in Fig.~\ref{fig:llm}(a)) and is instructed to generate a population of genotypes (shown in Fig.~\ref{fig:llm}(b)). These genotypes are then passed to a text-to-3D generative model $p_{\theta}$ for synthesizing the corresponding 3D designs.

\subsubsection{Offspring Generation} 
\begin{figure*}
    \centering
    \includegraphics[width=1.00\textwidth]{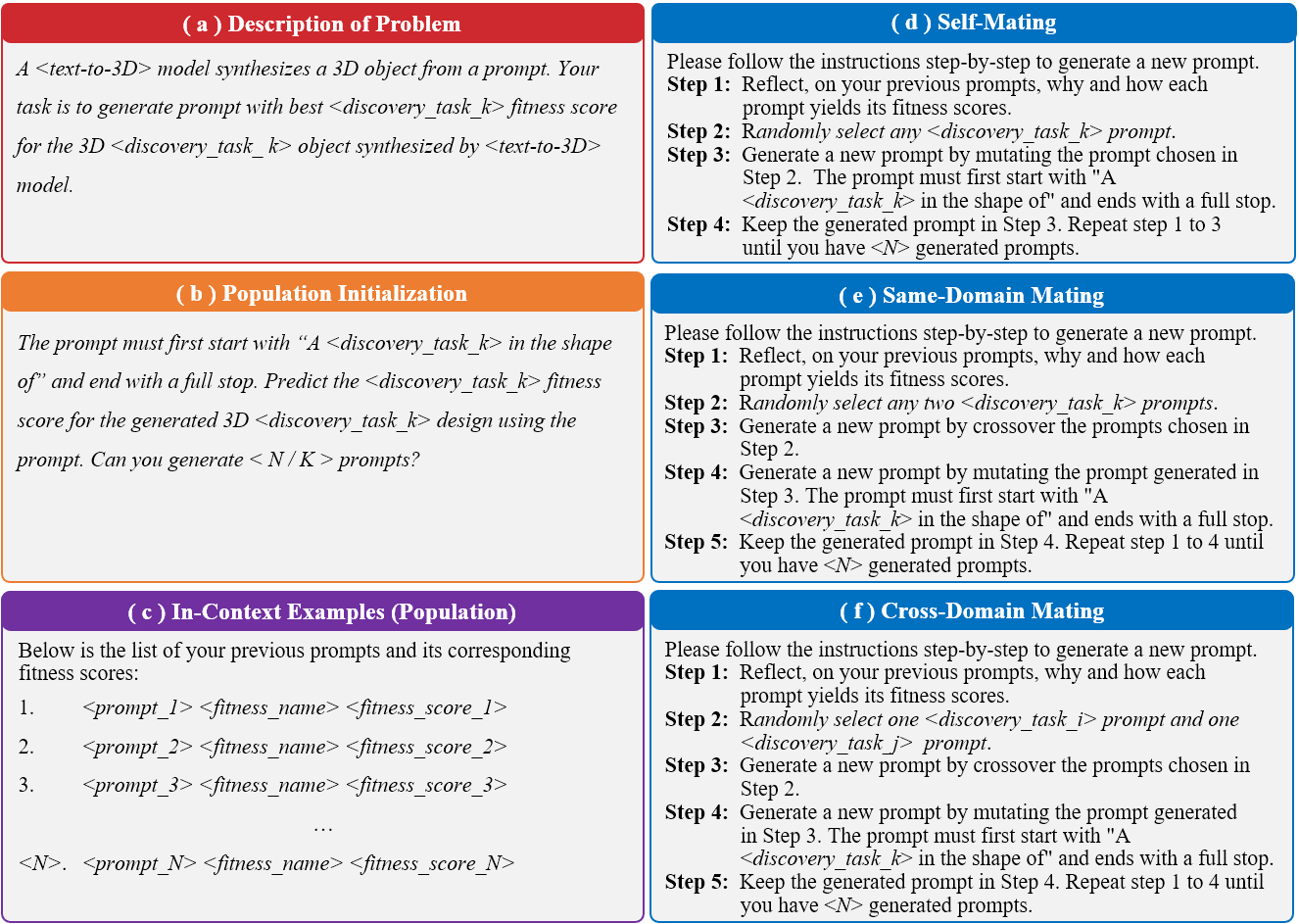}
    \caption{Instruction sets for the LLM to function as an evolutionary optimizer. LLM2TEA enables the LLM to initialize and evolve prompts guided by various physical and visual criteria in the evolution process.}
    \label{fig:llm}
\end{figure*}

It has been demonstrated that the in-context learning capability of LLM can incorporate historical objective evaluation data to enhance optimization guidance~\cite{liu2023LLMEvoOptimizers}. Building on this insight, the proposed LLM2TEA provides the LLM with the current genotype population and instructs it to reflect on them (as shown in Fig.~\ref{fig:llm}(c)) prior to generating offspring. This reflection ensures the context-aware execution of the assigned mating operations, aligning genetic variation effectively with the objectives of the algorithm.

During offspring generation, the LLM is instructed to select parent genotypes from the current population and undergo crossover and/or mutation operations. Conventional EAs typically employ simple genetic crossover operators~\cite{pavai2016survey} or probabilistic models, such as Gaussian distributions~\cite{ono1997real}, as mutation operators~\cite{hinterding1995gaussian}. However, these traditional crossover and mutation techniques exhibit two main limitations that could inhibit the effectiveness of prompt search techniques. Firstly, meaningful prompts in natural language often entail implicit constraints inherent in language sequences. Traditional crossover and mutation operators fail to account for these implicit constraints, potentially resulting in nonsensical prompts that may not effectively generate reasonable outputs. Secondly, considering that an LLM contains extensive prior knowledge, harnessing for crossover and mutation~\cite{meyerson2023language} can facilitate the generation of more coherent and contextually appropriate prompts by utilizing this embedded knowledge.
Building on the above considerations, LLM plays the role of an evolutionary operator to facilitate the mating process. This process involves selecting parents, followed by crossover and/or mutation operations to construct a genotype, and finally, generating a design conditioned by the constructed genotype. In LLM2TEA, several LLM-based self-mating, same-domain mating, and cross-domain mating operators are proposed, which are described as follows:
\begin{itemize}[leftmargin=1em]
    \item \textit{Self-Mating}: Instruction is provided to the LLM to select a parent genotype from the current population and construct a new genotype based on its context.
    \item \textit{Same-Domain Mating}: Instruction is provided to the LLM to randomly select two or more same-domain parent genotypes, constructing a new genotype by combining words from both. The LLM can mutate the new same-domain genotype further by randomly replacing some words.
    \item \textit{Cross-Domain Mating}: Instruction is provided to the LLM to randomly select two or more parent genotypes from different domains by combining words from both. The LLM can mutate the new cross-domain genotype further by randomly replacing some words.
\end{itemize}
During the mating process, if both selected parents belong to the same domain, the same-domain mating operator (shown in Fig.~\ref{fig:llm} (e)) kicks in; otherwise, a cross-domain mating operator (shown in Fig.~\ref{fig:llm} (f)) takes effect. Additionally, the self-mating operator (shown in Fig.~\ref{fig:llm} (d)) involving a single parent is triggered with a certain probability. The resulting genotypes are passed into a text-to-3D generative model, thereby synthesizing new 3D designs. The generated 3D design and its associated prompt are the designs of an offspring genotype. All the offspring genotypes will form the offspring population $O_{t}$ and are evaluated based on the physical and visual criteria. Subsequently, the offspring population is combined with the current population $P_{t}$ as the joint population. Any duplicate genotypes found in this joint population will be removed in the current generation population, retaining the unique genotype originating from the offspring population. The resulting joint population undergoes selection pressure based on the surviving genotypes.

\subsubsection{Selection} \label{sec:selection} A \emph{scalar fitness} measure to compare genotypes across domains is considered. Specifically, given $K$ tasks and $N$ genotypes in the joint population where $j \in \{1,...,N\}$, a set of \emph{factorial ranks} $\{r^{(1)}_{k},...,r^{(N)}_{k}\}$ for the $K$ tasks is determined by finding the ordered position of every genotype with respect to all other genotypes' \emph{factorial cost}. Here, this factorial cost is the genotype's fitness score. As such, each genotype in the joint population will have a set of factorial ranks for all tasks. Thereafter, the scalar fitness value $\varphi_{j}$ for the $j$-th genotype is computed as $\min_{k \in \{1,...,K\}} \{r_{k}^{(j)}\}$. The scalar fitness values of the joint population are used in any evolutionary selection technique to eliminate less fitting genotypes, allowing the surviving individuals to form the next generation of population in the evolutionary process.

\section{Experiments Setup}
The experimental setup to evaluate various LLM2TEA capabilities and the discovered designs is presented here. First, the configurations for various test scenarios are introduced. Next, the experimental settings of the LLM2TEA are elaborated.

\subsection{Test Scenarios}\label{sec:scenarios}
The performance of LLM2TEA is observed in the context of engineering design instances. Following Equations~\eqref{eq:obj} and~\eqref{eq:moo_obj}, the experiments are set up as a two-task design problem that targets to find novel car and airplane designs concurrently in a single EMT run. As such, the \emph{car} domain is $\textbf{t}_1 = \emph{\text{``A car in the picture''}}$, while the \emph{airplane} domain is $\textbf{t}_2 = \emph{\text{``An airplane in the picture''}}$. The effectiveness of LLM2TEA with ChatGPT 3.5~\cite{OpenAI} in driving evolution and Shape-E~\cite{Jun2023} as the auxiliary text-to-3D generative model is considered. The search space \(\mathbb{Z}^{L \times V}\) for effective prompts is thus \(L=77\) and \(V \approx 50K\) for Shape-E, while a maximum sequence context length of \(C \approx 16K\) and \(V_{c} \approx 50K\) for ChatGPT 3.5. In terms of $f_{visual}$ in Equation~\eqref{eq:obj} and \eqref{eq:moo_obj}, BLIP-2 VLM~\cite{li2023blip} is used as the evaluator to assess the visual performance of the generated designs. This visual criterion has a contribution weight of $\alpha = 0.55$, taking priority over the physical criteria in the objective. In addition, \(\underline{c} \geq 0.5\) and \(\bar{c} \leq 1.0\) in Equation~\eqref{eq:moo_obj} are set to constrain the exploration of generated designs within and near domain boundaries. Four test scenarios are considered to study the effectiveness of the LLM2TEA:
\begin{itemize}[leftmargin=1em]
    \item \textit{Test Scenario I}: This test scenario assesses the design diversity attained by LLM2TEA in a multi-objective setting that aims to minimize aerodynamic drag in both car and airplane designs, simultaneously maximizing aerodynamic lift in airplanes while minimizing aerodynamic lift in car designs.
    \item \textit{Test Scenario II}: This test scenario considers a minimization of the projected frontal area \cite{de2000computational} as $f_{phy}$.
    \item \textit{Test Scenario III}: This test scenario minimizes aerodynamic drag criterion.
    \item \textit{Test Scenario IV}: This test scenario assesses the creativity capability of LLM2TEA in discovering novel designs. Specifically, the two-task optimization tasks are defined as $\textbf{t}_{1}=$ ``\emph{banana car}'' and $\textbf{t}_{2}=$ ``\emph{banana airplane}''. The generated designs are grounded in real-world performance-based optimization criteria, allowing for an assessment of creativity while ensuring conformity to real-world requirements.
\end{itemize}
Aerodynamic drag and lift of generated designs are computed using OpenFOAM~\cite{weller1998tensorial} and normalized. OpenFOAM is an open-source computational fluid dynamics software toolkit primarily used for simulating fluid flow, heat transfer, turbulence, chemical reactions, and other related physical phenomena.

\begin{figure*}
    \centering
    \begin{tabular}{cc}
        \includegraphics[width=0.48\textwidth]{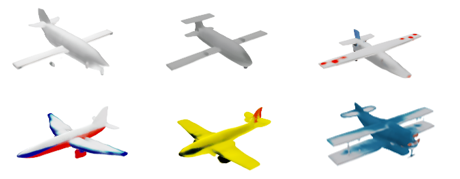} &
        \includegraphics[width=0.48\textwidth]{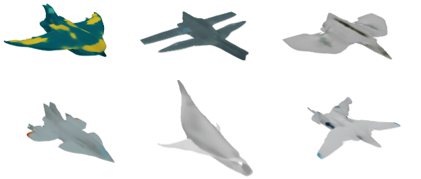} \\   
        \textbf{(a) Baseline (Airplanes)} & \textbf{(c) LLM2TEA (Airplanes)}\\ [5pt]
        \includegraphics[width=0.48\textwidth]{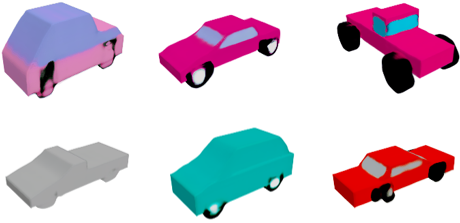} &
        \includegraphics[width=0.48\textwidth]{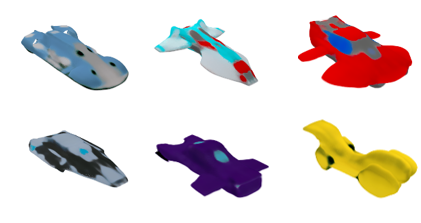} \\
        \textbf{(b) Baseline (Cars)} & \textbf{(d) LLM2TEA (Cars)} \\ [5pt]
    \end{tabular}
    \caption{Examples of designs generated by text-to-3D generative model are shown. In contrast to the vehicle body shapes generated directly by the model without the LLM2TEA (Fig.~\ref{fig:designs}(a) and (b)), the designs discovered with LLM2TEA deviate from these conventional shapes, since the model are guided to synthesize designs that not only satisfy physical criteria but also embody the visual characteristics of a car or an airplane, as illustrated by some of the representative example designs in Fig.~\ref{fig:designs}(c) and (d). The cross-domain search enabled by the multitask setting facilitates the discovery of hybrid vehicle designs, as most evident in Fig.~\ref{fig:designs}(d).}
    \label{fig:designs}
\end{figure*}

\subsection{Parameter Settings}
In all experiments, the population size is set to $N=20$, and consistent random seeds are used throughout the evolution, with a terminating criterion of up to \(20\) generations. This results in up to \(400\) designs being generated and evaluated by the text-to-3D generative model in a single run. The classic tournament selection algorithm \cite{miller1995genetic} is used to select the $N$ genotypes that form the next generation of designs. Each experiment run is executed on a single shared compute node equipped with Intel Xeon Silver 64 CPU cores clocked at 2.10 GHz, 128 GB of RAM, and three Nvidia Quadro GV100 GPUs (each with 32 GB of RAM), with both evolution and simulation run in parallel.

\subsection{Evaluating Designs} In the context of LLM2TEA, an important observation is its ability to generate designs that deviate from its conventional form. This capability is particularly evident in cross-domain design scenarios, where integrating functional principles and constraints from multiple domains often produces artifacts with fused structural or behavioral traits, resulting in novelty. However, in existing generative design research in literature~\cite{demirel2024human,di2021generative,martorelli2023strategies,furtado2024task,jaisawal2021generative,mountstephens2020progress,agkathidis2015generative}, there is no clear definition or standard for the concept of ``novel'' design. This paper defines design novelty by examining the distinction between ``conventional'' and ``unconventional'' shapes. Specifically, randomly sampling shapes from a generative model based on a known domain $\textbf{t}$ (e.g., a car or an airplane) is considered conventional, i.e., designs sampled as $\textbf{x} \sim p_{\theta}(\textbf{x}|\textbf{t})$. Designs with shapes deviating from this conventional set are considered novel designs.

To quantify novelty, a Novelty Score metric is proposed based on this idea. Using a VLM, a generated 3D design is compared with the visual bias of a target concept specified in text form. Specifically, the VLM is tasked to measure the similarity distance between the design and the task label (such as ``\emph{A car}'' or ``\emph{A airplane}''). This measurement can indicate whether the design is visually similar to a conventional concept of the vehicle task that VLM has already learned and whether the text-to-3D generative model can be synthesized. As such, \textit{Novelty Score} metric is introduced which is formulated as:
\begin{equation} \label{eq:inno}
    NS(\textbf{x}, \hat{\textbf{z}}_{k}) = g_{vlm}(\textbf{x}, \hat{\textbf{z}}_{k}) - \mathbb{E}_{p_{\theta}(\hat{\textbf{x}}|\hat{\textbf{z}}_{k})} \left[ g_{vlm}(\hat{\textbf{x}}, \hat{\textbf{z}}_{k}) \right], 
\end{equation}
where $\hat{\textbf{z}}_{k}$ is the $k$-th task label with the format ``A $< \textbf{t}_{k} >$'' and $g_{vlm}(\cdot)$ is the independent VLM (e.g., CLIP\cite{pmlr-v139-radford21a}) to measure the similarity distance between a design and its corresponding task label. This VLM is used in the Shape-E model to train and generate 3D designs, aligning with the visual bias inherent in the text-to-3D generative model. Notice that when $NS(\cdot) \leq 0$, the design is considered to exhibit only conventional features. Conversely, when $NS(\cdot) > 0$, the design is considered to be ``\emph{novel}'', indicating a deviation from the conventional forms known to the VLM. On a separate note, Monte Carlo estimation is adopted here to approximate $\mathbb{E}_{p_{\theta}(\hat{\textbf{x}}|\hat{\textbf{z}}_{i})} \left[ g_{vlm}(\hat{\textbf{x}}, \hat{\textbf{z}}_{i}) \right]$ using \(1,000\) generated samples per task as the baseline mean, taking into account the variety of conventional shapes that resemble the target task.

\section{Results and Discussions}
This section presents the experiment results in the following order. First, the designs discovered during the experiments are showcased. Next, the diversity of these designs is analyzed, followed by a discussion of the various components of the LLM2TEA to evaluate its capabilities and effectiveness. Finally, experiments demonstrating the 3D printing of the generated designs are presented.

\subsection{Results of Discovered Designs}
Some of the most representative examples from all test scenarios of vehicle designs uncovered in the text-to-3D generative model, Shape-E, are presented in Fig.~\ref{fig:designs}. Vehicle designs from the baseline exhibit conventional vehicular features, as shown in Fig.~\ref{fig:designs}(a) and (b), and have a distinct contrast with the novel designs found with the LLM2TEA, as illustrated in Fig.~\ref{fig:designs}(c) and (d). Such novel designs that deviate from conventional form result from guiding the generative models to synthesize complex prompts. These prompts lead to designs that not only satisfy the physical criterion but also fulfill the visual specification of a car or an airplane. A clear distinction can be observed between designs generated with LLM2TEA  and those from the baseline model. The cross-domain knowledge learning capability of LLM2TEA enables the genetic material from parent prompts across different domains to undergo crossover and mutation, thereby facilitating the transfer of knowledge across domains. This process generates offspring prompts that drive the discovery of hybrid vehicle designs. These designs feature unconventional body shapes and structures, such as banana-like curvatures or extended rear spoilers, challenging traditional car and airplane design concepts.

\textit{\textbf{Remark}}: \textit{In LLM2TEA, the implicit genetic transfer occurring during cross-domain mating facilitates the creation of hybrid designs that combine the distinctive traits of each parent genotype from different domains. On a separate note, LLM and text-to-\emph{X} generative models known for its hallucination phenomenon~\cite{maleki2024ai} may lead to synthesizing outputs that deviate from the factual or the input prompt. Such a hallucination outcome is not always detrimental but can foster positive creativity~\cite{jiang2024survey}. Both of these algorithmic properties create the opportunity for emerging design trends, facilitating the discovery of novel designs.}

\subsection{Diversity of Designs} 
\begin{figure*}
    \centering
    \includegraphics[width=1.00\textwidth]{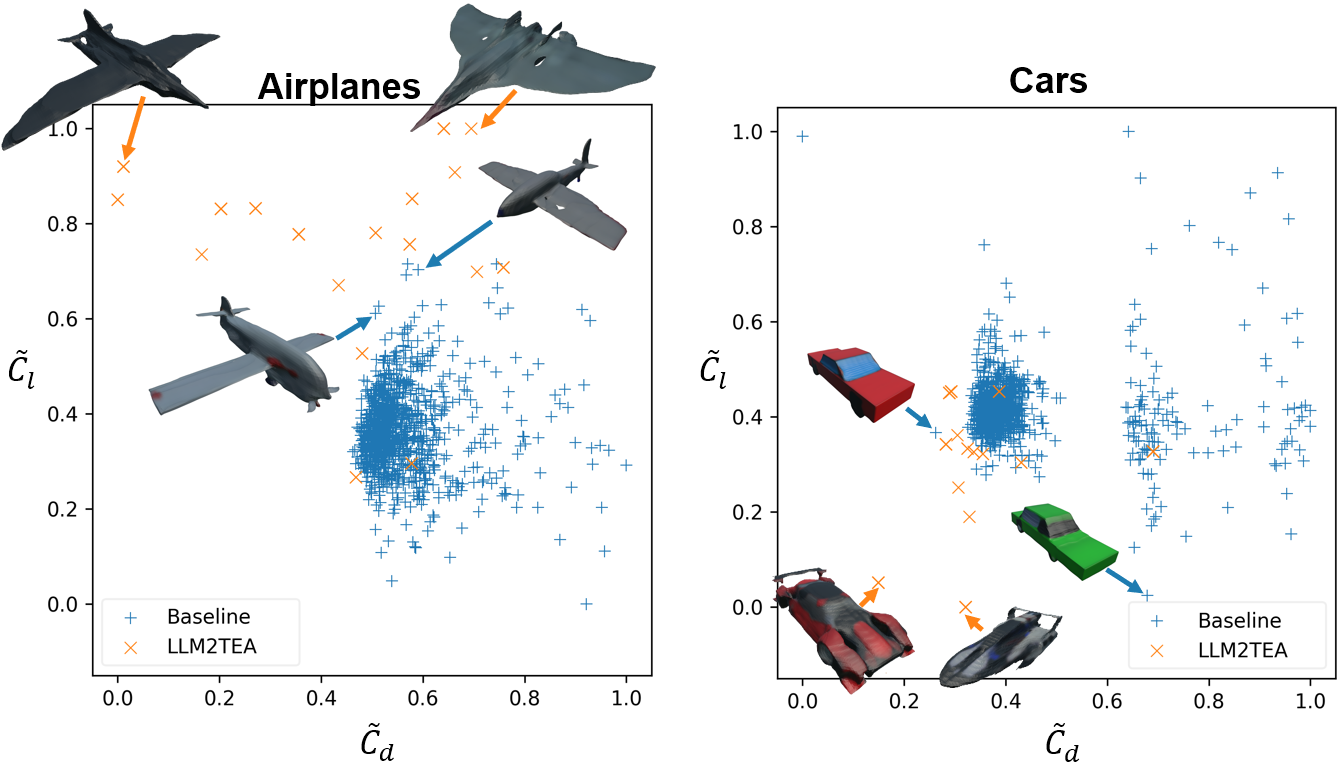}
    \caption{Test Scenario I compares between baseline designs generated by the baseline text-to-3D generative model without LLM2TEA to designs discovered with LLM2TEA during the last five generations of a single evolutionary run. The designs discovered with LLM2TEA demonstrated superior expected aerodynamic performance (exhibiting higher normalized lift ($\tilde{C}_{l}$) and lower normalized drag ($\tilde{C}_{d}$) scores for airplanes, as well as lower $\tilde{C}_{l}$ and $\tilde{C}_{d}$ scores for cars) and are visually more novel than the baseline designs. (Note that the illustrations presented here are not drawn to scale.)}
    \label{fig:mtmo_compare}
\end{figure*}

In Test Scenario I, the best-performing designs from the last five generations in a multi-objective setting of a single run using LLM2TEA are compared against baseline designs generated with a text-to-3D generative model without LLM2TEA. As shown in Fig.~\ref{fig:mtmo_compare}, the designs discovered with the LLM2TEA exhibit superior aerodynamic performance compared to the baseline. Specifically, airplane designs demonstrated better lift ($\tilde{C}_l$) and drag ($\tilde{C}_d$) than baseline designs, with similar trends observed for car designs. Apart from that, the baseline text-to-3D generative model tends to discover designs clustered in localized regions, whereas LLM2TEA produces more diverse designs, as evident from the visual inspection of the generated airplane designs. To quantitatively assess the diversity, Table~\ref{tbl:innohv} presents the hypervolume indicator~\cite{ishibuchi2018specify} measured over the physics objectives ($\tilde{C}_{l}$ and $\tilde{C}_{d}$) space, which ranges from \(0.0\) to \(1.0\). LLM2TEA achieves hypervolume values \(2.74\) greater for airplane designs and \(1.97\) times for car designs compared to the baseline. These results demonstrated that the proposed LLM2TEA effectively explores design spaces to uncover novel designs with strong physics performance.

\subsection{Effectiveness of LLM2TEA}
A series of experiments was conducted to objectively examine the capabilities of LLM2TEA to discover novel solutions. The findings are presented here.

\begin{figure*}
    \centering
    \begin{tabular}{m{9cm} m{8cm}}
    \includegraphics[width=0.48\textwidth]{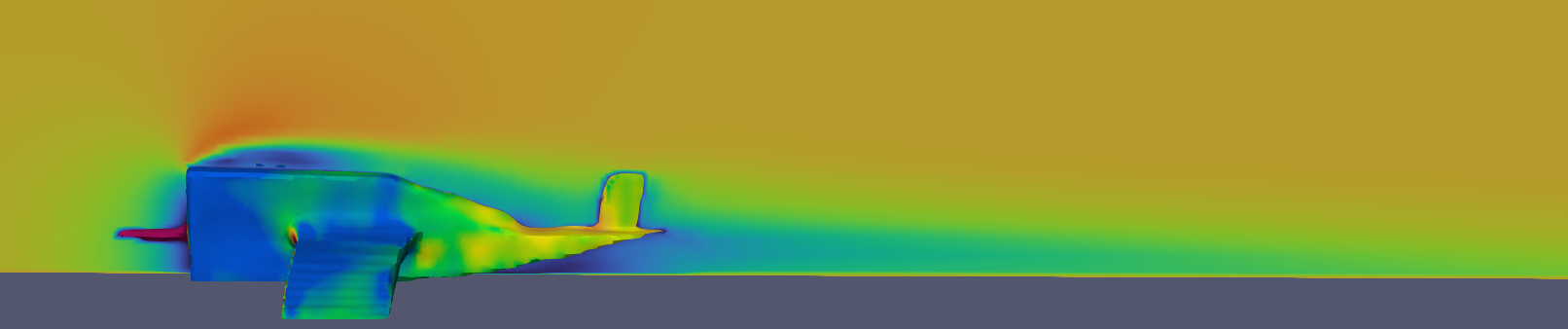} & \makecell[c]{``A airplane in the shape of a mysterious shadow.''} \\ [30pt]
    \includegraphics[width=0.48\textwidth]{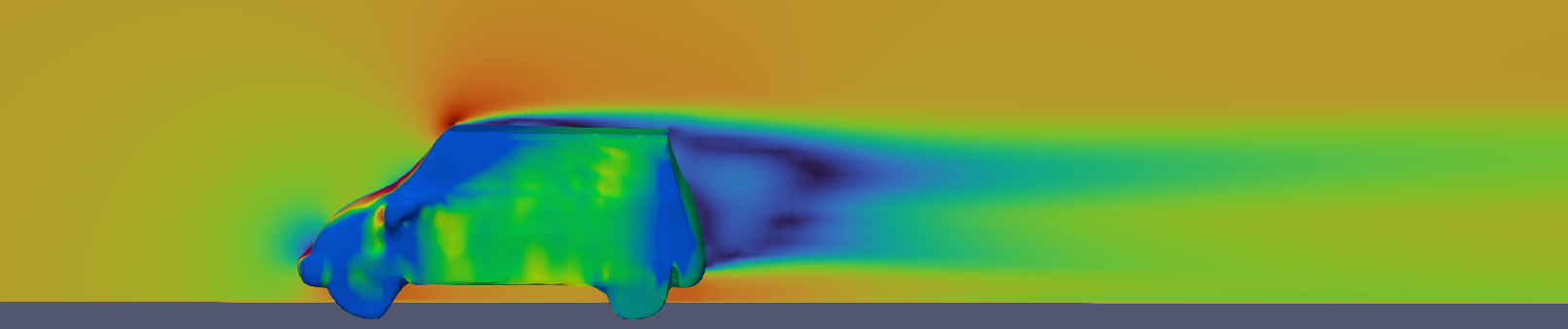} & \makecell[c]{``A car in the shape of a compact city hatchback with a \\ practical design and efficient fuel consumption.''} \\ [30pt] 
    \multicolumn{2}{c}{\textbf{(a) LLM2TEA with Visual Criteria Only}} \\ [5pt]
    \includegraphics[width=0.48\textwidth]{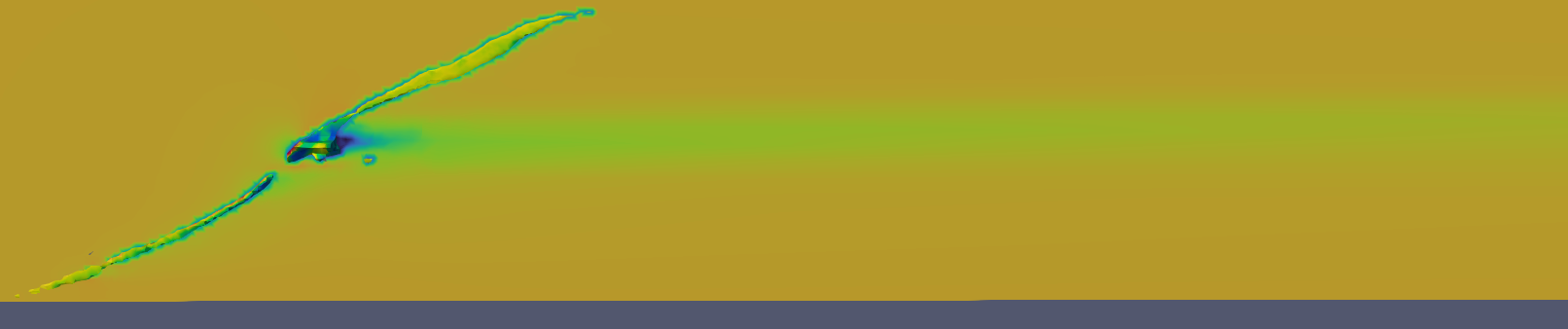} & \makecell[c]{``A airplane in the shape of a slender arrow piercing \\ gracefully through the fog.''} \\ [30pt]
    \includegraphics[width=0.48\textwidth]{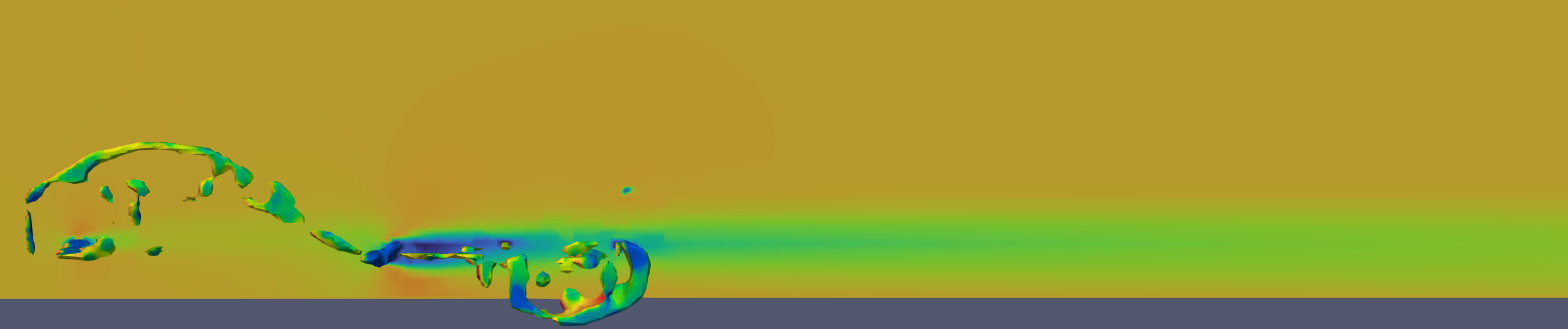} & \makecell[c]{``A car in the shape of branched chain.''} \\ [30pt]
    \multicolumn{2}{c}{\textbf{(b) LLM2TEA with Physical Criteria Only}} \\ [5pt]
    \includegraphics[width=0.48\textwidth]{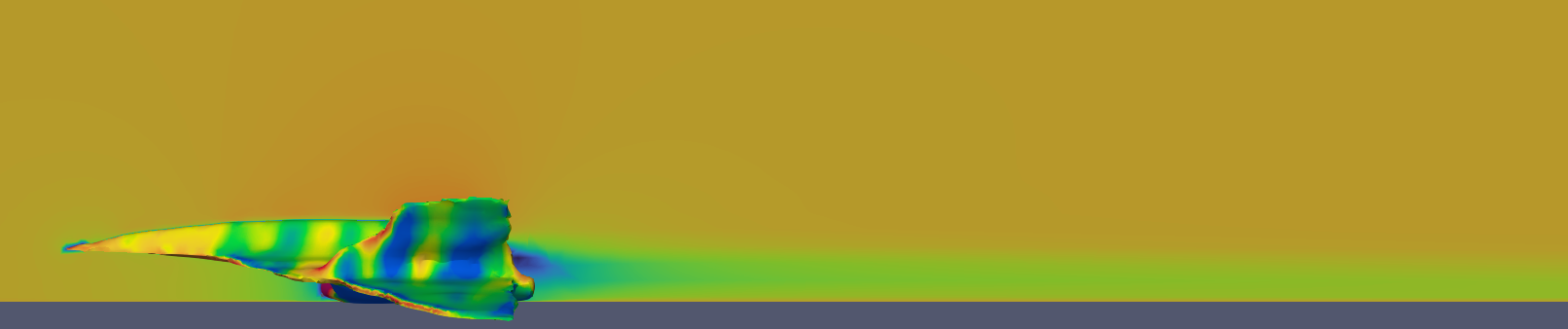} & \makecell[c]{``A airplane in the shape of a modern cutting-edge canard \\ stealth jet blending state-of-the-art luxury technology \\ with advanced high-speed fuselage elements for \\ exceptional drag reduction and efficiency.''} \\ [30pt]
    \includegraphics[width=0.48\textwidth]{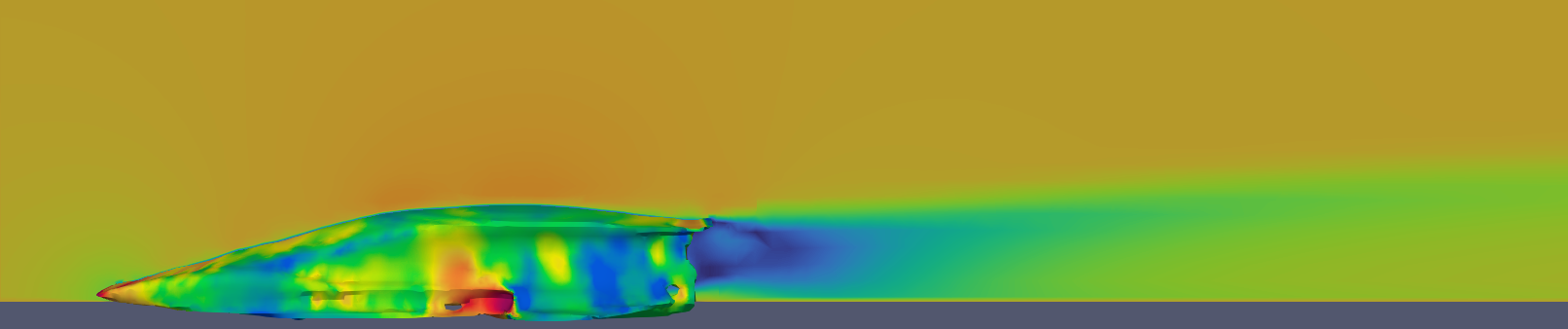} & \makecell[c]{``A car in the shape of a modern compact sedan \\ with a sleek profile and integrated LED headlights.''} \\ [30pt] 
    \multicolumn{2}{c}{\textbf{(c) LLM2TEA with Physical \& Visual Criteria}} \\ [5pt]
    \includegraphics[width=0.48\textwidth]{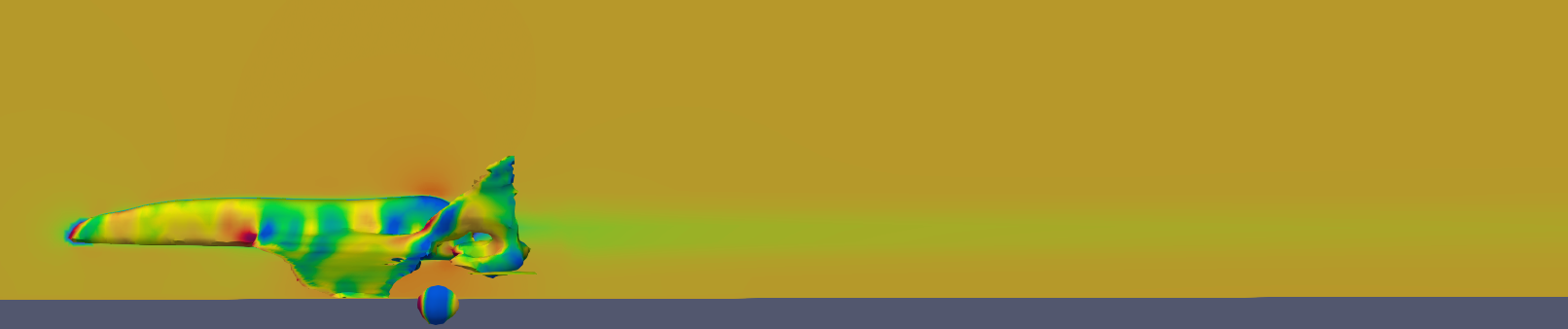} & \makecell[c]{\textbf{Predefined Prompt:} \\``An airplane''} \\ [30pt]
    \includegraphics[width=0.48\textwidth]{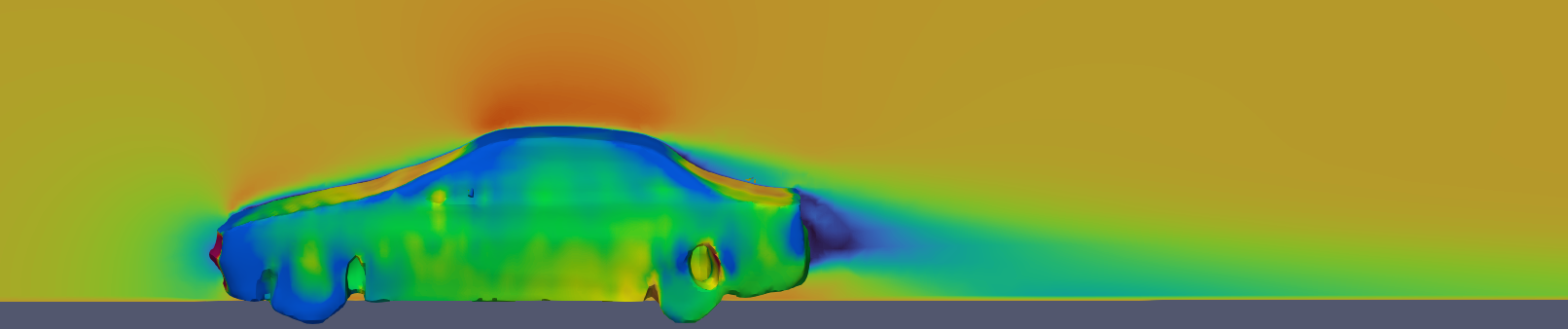} & \makecell[c]{\textbf{Predefined Prompt:} \\ ``A car''} \\ [30pt] 
    \multicolumn{2}{c}{\textbf{(d) Baseline}} \\ [5pt]
    \end{tabular}
    \caption{Examples of text prompts (right) used to generate the designs (left) are shown, with colors indicating the magnitude of the fluid velocity, where brighter colors correspond to higher velocity. When LLM2TEA evaluates using only visual criteria, the resulting vehicles tend to have rectangular shapes and high roofs, which correspond to poor drag performance, as illustrated in Fig.~\ref {fig:scientific} (a). Conversely, relying solely on the physical criterion will degrade the search performance with more highly fragmented vehicle designs discovered, such as those shown in Fig.~\ref{fig:scientific} (b). As such, by considering both physical and visual criteria simultaneously, LLM2TEA can discover practical designs (Fig.~\ref{fig:scientific} (c)) featuring body shapes that yield better drag performance compared to baseline (Fig.~\ref{fig:scientific} (d)).}
    \label{fig:scientific}
\end{figure*}
\renewcommand{\arraystretch}{1.5}
\begin{table}
    \centering
	\caption{Comparison of Hypervolume between Baseline and LLM2TEA in Single Evolutionary Run}
	\begin{tabular}{lcc}
		\hline
        Method & Airplane Task & Car Task \\
		\hline
        Baseline & $0.3111$ & $0.4907$ \\
        LLM2TEA & $\mathbf{0.8515}$ & $\mathbf{0.9702}$ \\
        \hline
	\end{tabular}
	\label{tbl:innohv}
\end{table}

\subsubsection{Effect of Evaluation Criteria}
This subsection examines the effects of the objective functions defined in Section~\ref{sec:problem_setting}, demonstrating their ability to guide LLM2TEA toward finding the best-performing designs. As shown in Fig.~\ref{fig:scientific}, when guided solely by visual criteria, LLM2TEA generates vehicle designs with wide body shapes and high roofs that retain the characteristics of a car or airplane but exhibit poor aerodynamic performance (Fig.~\ref{fig:scientific}(a)). In contrast, designs evaluated solely on physical criteria will result in the production of more highly fragmented designs, such as the example presented in Fig.~\ref{fig:scientific}(b). Therefore, incorporating both physical and visual criteria enables LLM2TEA to discover practical designs with body shapes that achieve the lowest drag (Fig.~\ref{fig:scientific}(d)) compared to those in Fig.~\ref{fig:scientific}(a) through (c).  Notably, over \(73\%\) of the generated vehicle designs outperform the top \(1\%\) designs produced by the text-to-3D baseline in terms of physical performance. These results suggest that LLM2TEA effectively uncovers novel yet practical designs.

\subsubsection{Evolution Trends of LLM2TEA}
\begin{figure*}
    \centering
    \includegraphics[width=1.00\textwidth]{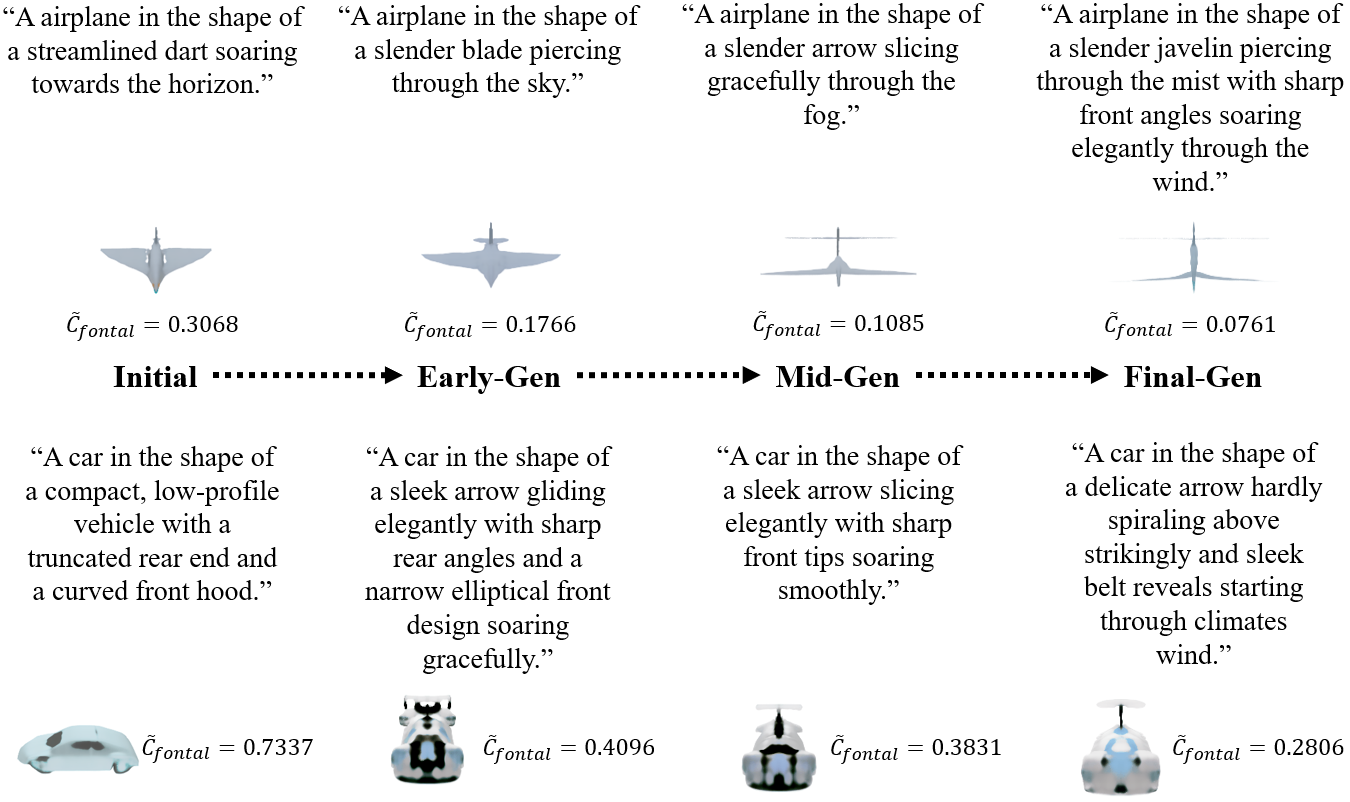}
    \caption{Example of the evolution of prompts and their corresponding vehicle designs during an LLM2TEA evolutionary run in Test Scenario II, which aims to minimize the projected frontal area. Note that the technical indicator (\(\tilde{C}_{frontal}\)) provided is a computed value from the evaluator and does not reflect the actual real-world geometry.}
    \label{fig:evofrontal}
\end{figure*}

\begin{figure*}
    \centering
    \includegraphics[width=1.00\textwidth]{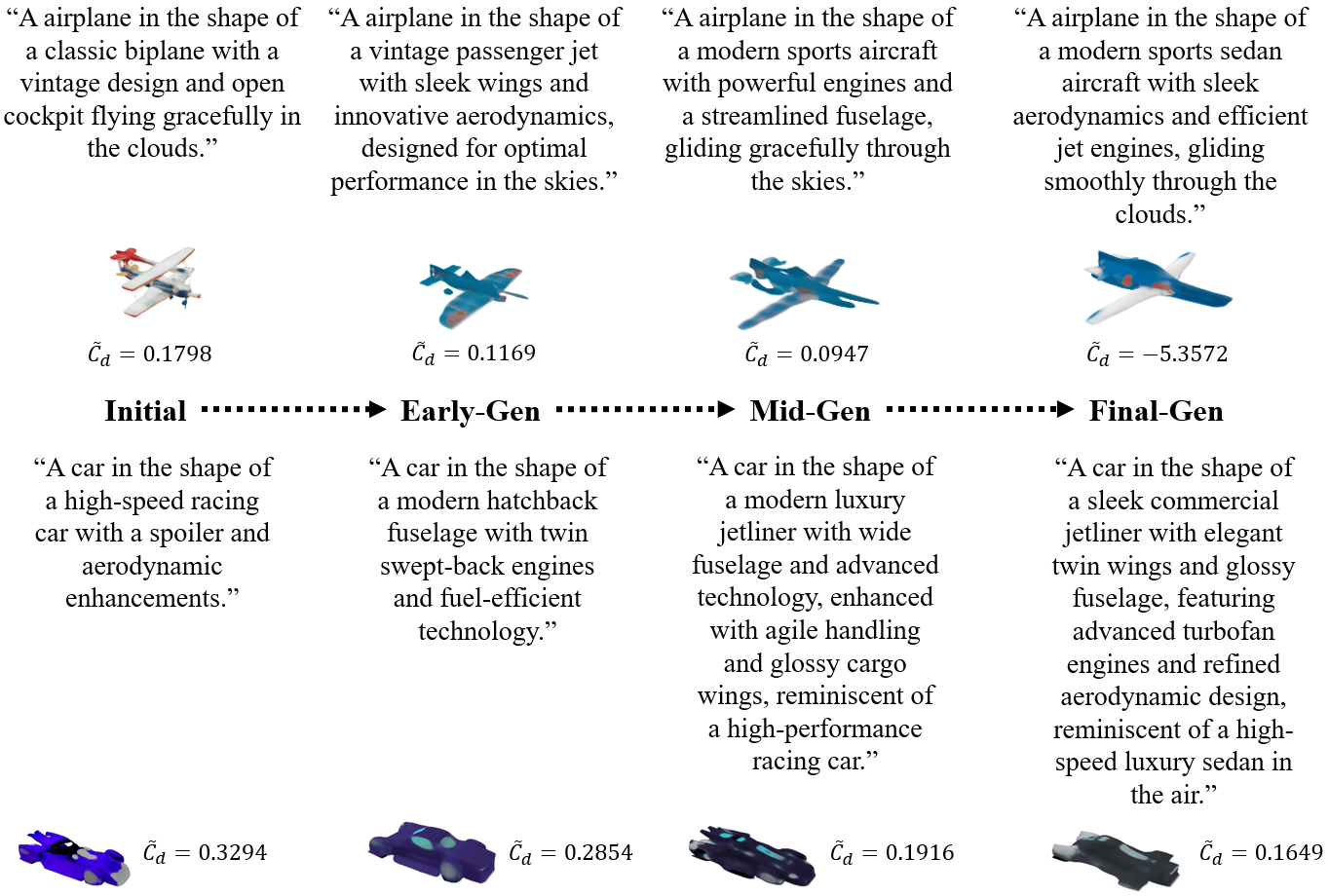}
    \caption{Example of the evolution of the prompts and its corresponding vehicle designs during an LLM2TEA evolutionary run in Test Scenario III, which aims to minimize drag. Note that the technical indicator (\(\tilde{C}_{d}\)) provided is a computed value from computational physics simulation and does not reflect the true real-world performance.}
    \label{fig:evodrag}
\end{figure*}

\begin{figure*}
    \centering
    \includegraphics[width=1.00\textwidth]{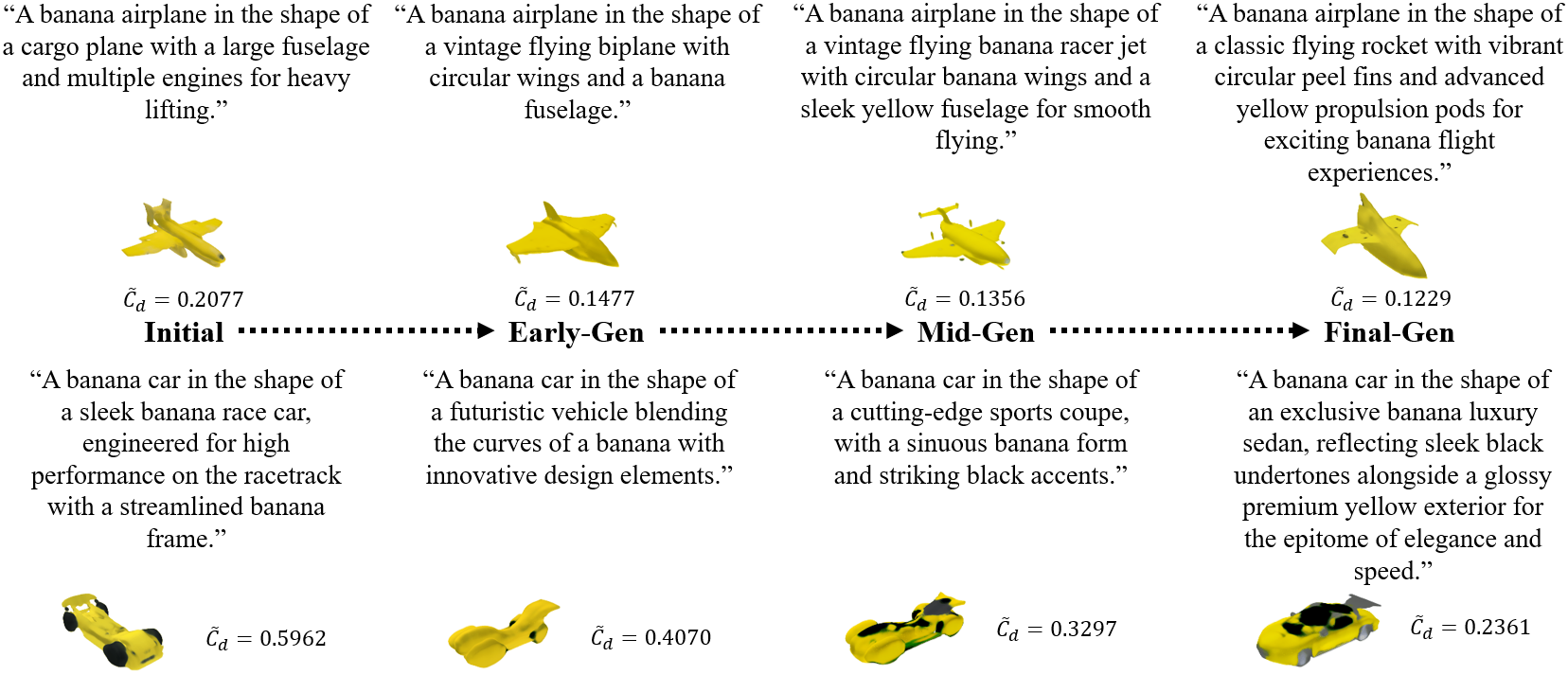}
    \caption{Example of the evolution of the creative prompts and its corresponding creative vehicle designs during an LLM2TEA evolutionary run in Test Scenario IV, which aims to minimize the drag. Note that the technical indicator (\(\tilde{C}_{d}\)) provided is a computed value from computational physics simulation and does not reflect the true real-world performance.}
    \label{fig:evocreative}
\end{figure*}

Table \ref{tbl:fitness} presents LLM2TEA average fitness scores across generations in the single evolutionary run. A distinct trend of gradual improvements in the initial phases can be observed, particularly for the airplane design task. This progression becomes more pronounced in the latter half of the run, indicating that the evolutionary search led to the discovery of better aerodynamic designs. Further examination of the results in all test scenarios supports the observed trend (Figs.~\ref{fig:evofrontal}, ~\ref{fig:evodrag}, and ~\ref{fig:evocreative}). While the initial vehicle designs exhibit body shapes resembling conventional forms, LLM2TEA enables the designs to evolve and deviate from their conventional forms, with more significant changes observed from the mid-generation phase of the evolutionary search.

\renewcommand{\arraystretch}{1.5}
\begin{table*}
    \centering
	\caption{Population Mean Fitness Score and its Variance Across Generations in Single Evolutionary Run}
	\begin{tabular}{lccccc}
		\hline
		Task & Initial & Gen. 5 & Gen. 10 & Gen. 15 & (Final) Gen. 20 \\
		\hline
		Airplane & $0.2683 \pm 0.0012$ & $0.1596 \pm 0.0015$ & $0.1485 \pm 0.0014$ & $0.0662 \pm 0.0415$ & $-0.0216 \pm 0.0427$ \\
        Car & $0.0981 \pm 0.0001$ & $0.0609 \pm 0.0001$ & $-0.0531 \pm 0.0514$ & $-0.0336 \pm 0.0343$ & $-0.1339 \pm 0.1388$ \\
        \hline
	\end{tabular}
    \vspace{3em}
	\label{tbl:fitness}
\end{table*}
\begin{figure*}
    \centering
    \includegraphics[width=1.00\textwidth]{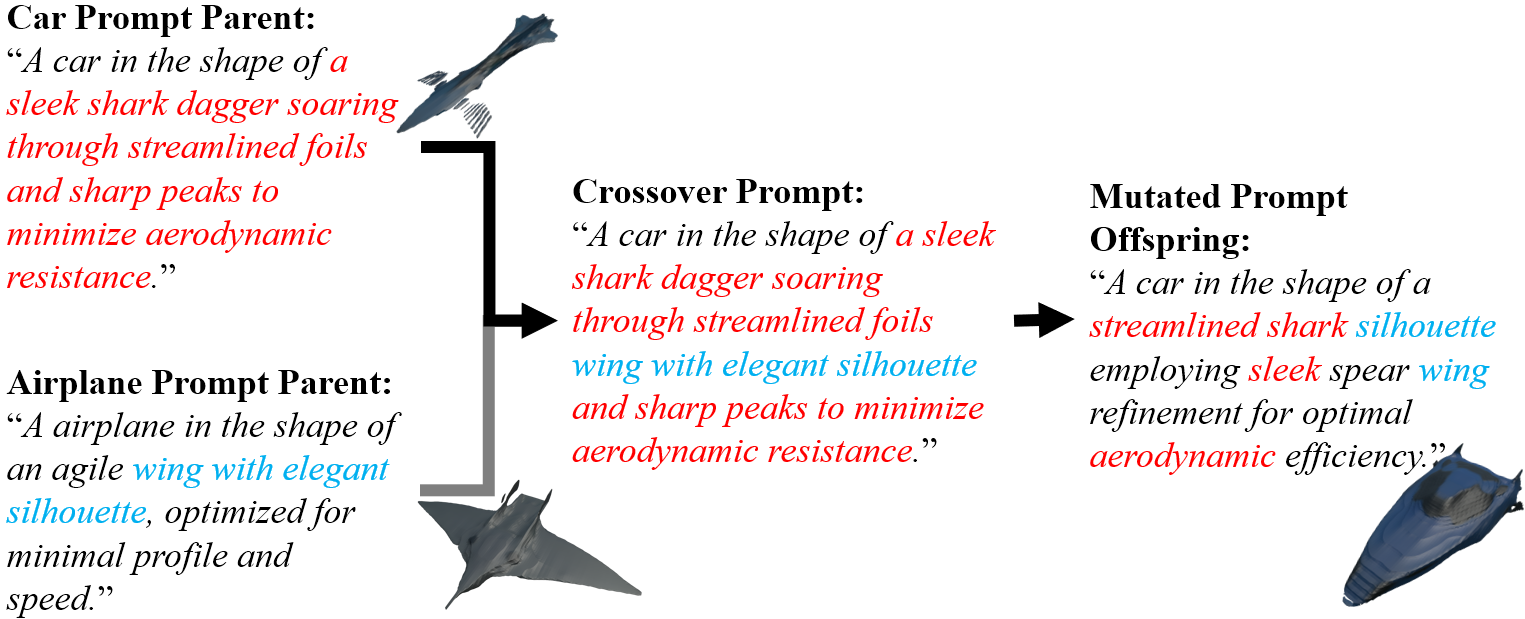}
    \caption{Illustration of prompt crossbreeding between \textcolor{red}{car} and \textcolor{cyan}{airplane} domains, demonstrating multitask adaptation in the evolution of LLM2TEA.}
    \label{fig:transfer}
\end{figure*}
\begin{figure*}
    \centering
    \begin{tabular}{ccc}
     \includegraphics[width=0.31\textwidth]{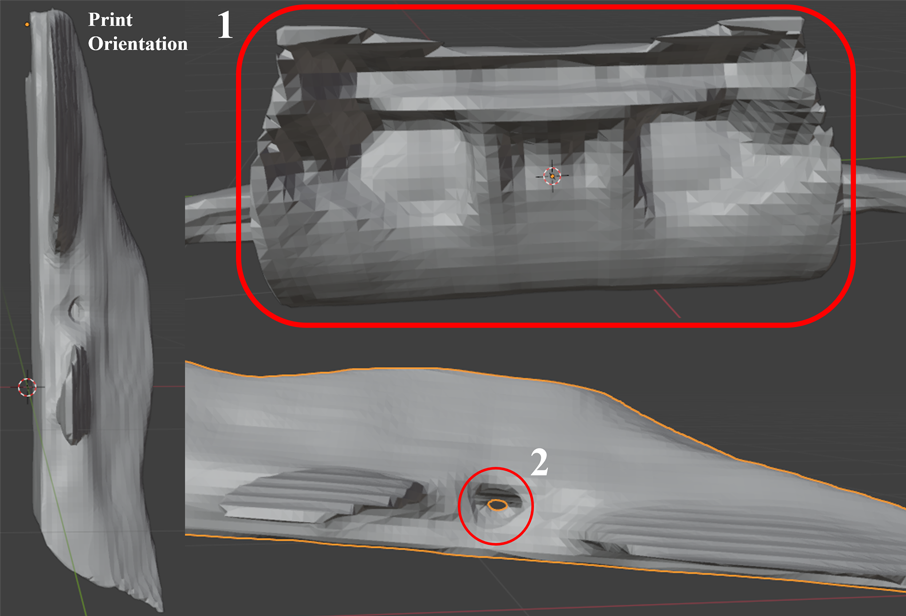} &
     \includegraphics[width=0.31\textwidth]{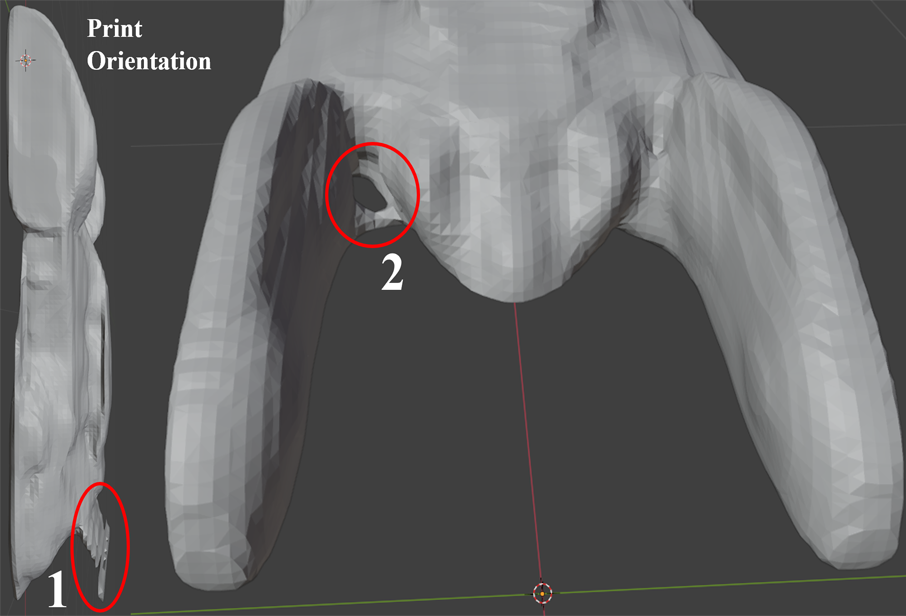} &
     \includegraphics[width=0.31\textwidth]{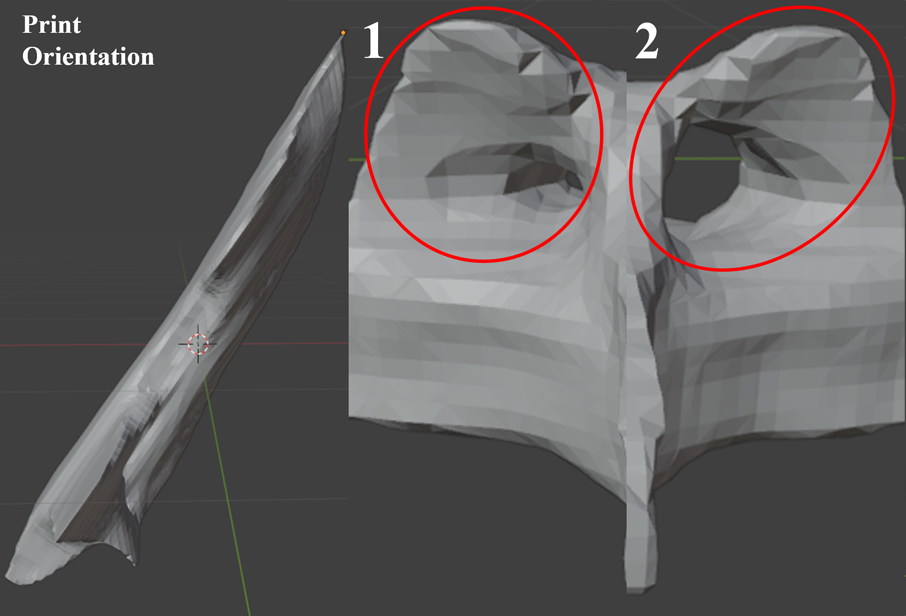} \\
     \includegraphics[width=0.31\textwidth]{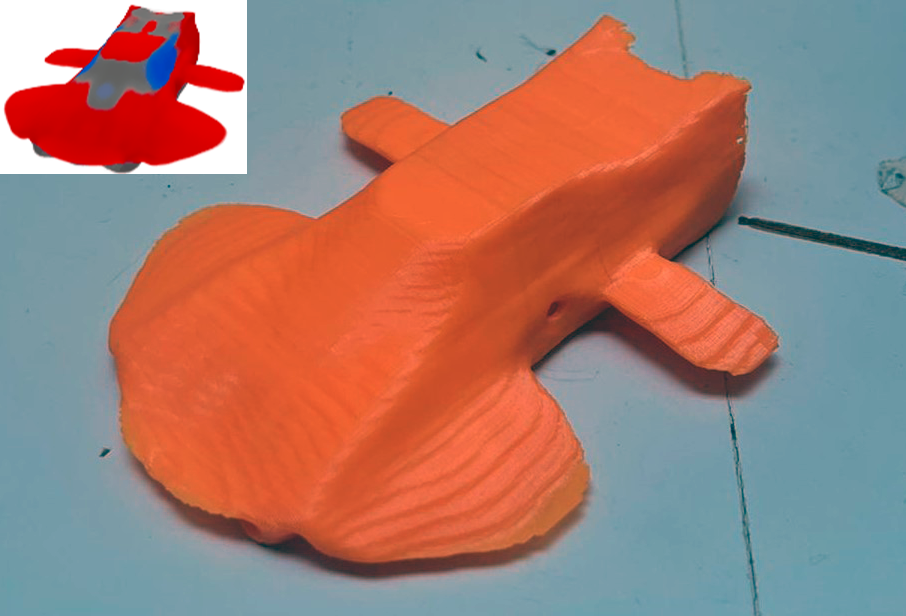} &
     \includegraphics[width=0.31\textwidth]{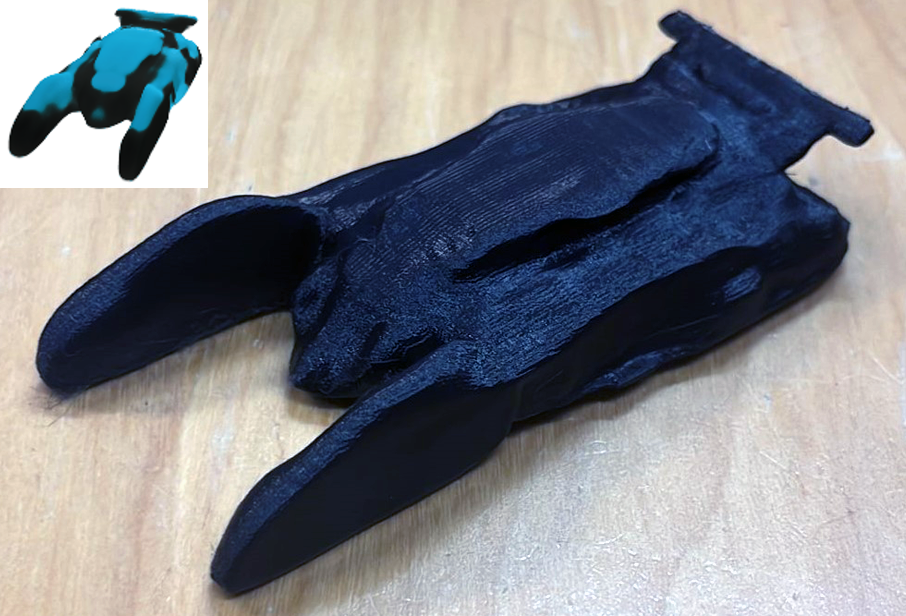} &
     \includegraphics[width=0.31\textwidth]{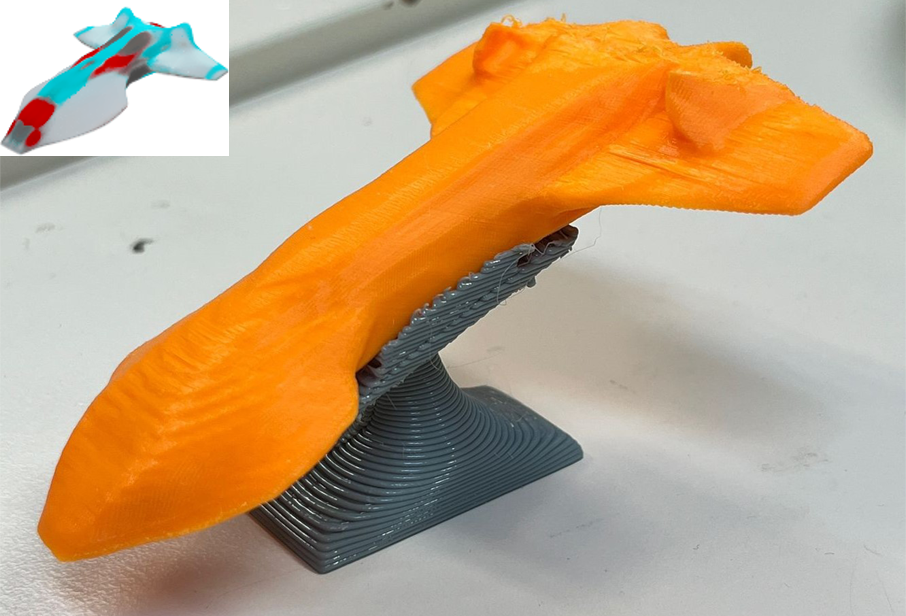} \\ 
     (a) & (b) & (c) \\ [3pt]
    \end{tabular}
    \caption{Study conducted on 3D printability of discovered designs found that the majority of designs generated by text-to-3D generative model carry some degree of defects, as illustrated in the top row of this figure. Some defects are easily repairable without losing the novelty of the design shapes by thickening the weak or thin structures, filling holes, and strengthening the internal cavity. Furthermore, the vehicle design's non-flat surfaces require printing orientation in a vertical position with various printing supports. The resulting 3D-printed designs, as shown in the bottom row, match closely with the original designs. This study demonstrated the potential use of GenAI for discovery and fast design prototyping, opening up new avenues for research and innovation.}
    \label{fig:3dprint}
\end{figure*}

\subsubsection{Behavior of Mating Operations} In Fig. \ref{fig:evodrag}, notice the less frequently used words included in the car prompts, such as ``\emph{wings}'' and ``\emph{jet}''. Such prompt constructs are the result of the LLM2TEA mating operations. Self-mating and same-domain mating operations construct same-domain prompts that utilize commonly used terms, resulting in vehicle designs with conventional body shapes. In contrast, cross-domain mating operation, with its cross-concept knowledge integration, enables the crossbreeding between different domains by combining shape-defining words from these domains to produce a new genotype. The resulting phenotype generated with this new genotype exhibits unconventional body shapes with a hybrid appearance, combining distinctive features of these domains while still satisfying the physical criteria (Fig.~\ref{fig:transfer}).

Since the LLM2TEA cross-concept knowledge transfer mechanism operates by transferring representative words from one task to another, the effectiveness of this capability can be observed by examining the percentage overlap between each task's vocabulary set. The vocabulary set for each task comprises representative word tokens of all the prompts generated throughout the evolutionary run. For a fair comparison, frequently occurring stopwords (such as ``\emph{a}'', ``\emph{the}'', ``\emph{is}'', etc) are pruned from the vocabulary set. Furthermore, all word tokens are reduced to its base word form by transforming the morphological variants to the base word. The resultant vocabulary set comprises the word tokens most representative of each task. Henceforth, an \(38.38\%\) overlap in the vocabulary sets between tasks, indicating a proportion of word tokens shared between tasks. The results demonstrated the effectiveness of the cross-domain knowledge-driven approach, which transfers a significant amount of words between tasks with LLM2TEA.

\subsubsection{Efficiency and Scalability of LLM2TEA} \label{sec:efficiency}
From our observations of the computational time of each key component in LLM2TEA, the computation time of the operators is negligible as proprietary LLMs is used, which runs on third-party servers. However, \(72.7\%\) of one generation total time is used in the OpenFoam physics simulation, which is typical in engineering design optimization problems. The remaining time is used for visual evaluations, accounting for \(20.5\%\) of the time, and text-to-3D generation utilizes the remaining \(6.8\%\) of the time. Similar to existing engineering design methods, where scaling up for bigger problems requires parallelizing physical simulation computations, LLM2TEA efficiently uses multicore CPU processing to compute the physical performance of generated designs. However, in contrast to existing evolutionary methods that require a large population size to yield good results, LLM2TEA can achieve good results with a small population size. Additionally, since the generated designs are available in digital 3D formats, the designs can be scaled down to fit into a unit cube, further reducing the simulation computation requirements.

\subsubsection{Quantification of Novel Solutions}
The effectiveness of LLM2TEA in discovering novel solutions is examined from a visual perspective using the proposed Innovation Score metric defined in Equation~\eqref{eq:inno}. Results indicate that \(69.12\%\) of airplane designs and \(46.23\%\) of car designs generated by LLM2TEA exceed the computed mean baseline, underscoring the efficacy of the multitask evolutionary process in guiding generative models toward the discovery of cross-domain novel designs. This further validates LLM2TEA’s capability to yield unconventional solutions not only across domains but also within a single domain, as illustrated in Fig.~\ref{fig:designs}(c) and (d), where the resulting designs exhibit more streamlined, futuristic, and sleeker body shapes that deviate from conventional norms.

\subsection{3D Design Printing}
The designs discovered with LLM2TEA can be 3D printed. Since the text-to-3D generative model (Shape-E) tends to generate designs that are non-conforming to manufacture in real-world conditions, most designs uncovered in the text-to-3D generative model possessed some degree of defects and fragmentation. For instance, as shown in the top row of Fig. \ref{fig:3dprint}, common defects, such as rough surfaces, sharp edges, open holes, and weak or thin structures, especially within internal cavities, can be easily resolved with minor repairs and infills to the designs before 3D printing. On the other hand, the non-flat and intricate surface details require printing orientation in a vertical position with minimal breakaway or tree-like printing supports. Additionally, the 3D designs require appropriate scaling to preserve its detailed geometric features during 3D printing.

Taking these factors into account, off-the-shelf SLA 3D printing is used due to its ability to retain such intricate details and its suitability for small-quantity prototyping. The resulting 3D-printed designs are shown in the bottom row of Fig.~\ref{fig:3dprint}, closely matching the digital visualization of the designs. The conceptual design process can be accomplished within weeks, from discovery to 3D printing the designs, with the start of examining the designs during a single evolutionary run in a matter of hours. This demonstration highlights the capability of LLM2TEA to transform AI-generated design outputs into tangible physical designs, underscoring its potential applications in fast design prototyping.

\textbf{Concluding Remark:} \label{sec:concluding} \emph{We hope that the insights presented in this experiment discussion section will provide researchers a better understanding of the challenges in the problem of interest and address the identified technical generation issues in LLMs and GenAI models using our proposed strategies. There are several ways practitioners can set up LLM2TEA to suit their specific needs, including defining the task domains with the variable \(\textbf{t}_{k}\), choosing a preferred physics simulator, and setting the hard constraint bound values \(\underline{c}\) and \(\bar{c}\) to control the exploration behavior. We learned that it is crucial to identify task domains that are sufficiently general, exhibit positive transfer of geometric characteristics between tasks, and integrate well with automatic mesh generation and simulation. Moreover, generated cross-domain designs should maintain validity for evaluations as exploration goes deeper into out-of-domain regions. In summary, the proposed method applies to any applications in science and engineering as long as the generated designs can be evaluated. These insights hold broader relevance to research communities, such as those in product design or animation, and other fields seeking to leverage LLM2TEA, LLMs, and GenAI models for novel design discovery.}

\section{Conclusion and Outlook}
This paper presents LLM2TEA, the first agentic AI designer operating with GEM, leveraging LLMs to explore and discover novel, physically viable designs across multiple domains. By integrating LLM-guided evolutionary search, text-to-3D generation, semantic evaluation, and physics simulation, LLM2TEA empowers designers to navigate vast, cross-domain design spaces, unlocking novel solutions that ground aesthetic novelty with real-world practicality. In particular, LLM2TEA is an agentic AI designer capable of expanding the boundaries of conventional design by leveraging knowledge across multiple domains. It explores a broader design space that extends beyond individual areas and embraces their overlap through the paradigm of evolutionary multitasking. This approach promotes the discovery of novel cross-domain designs through effective implicit knowledge transfer. Experimental results demonstrate significant improvements in design diversity (up to \(174\%\)) and physical performance (with over \(73\%\) surpassing leading baselines). The successful 3D printing of several generated designs further validates the real-world feasibility of LLM2TEA, underscoring its transformative potential for AI-driven, cross-domain innovation and practical design optimization.

While the empirical studies of the prototypical design optimization problem are an abstraction of real-world engineering design scenarios, the potential of the LLM2TEA in accelerating the conceptual design process and enhancing the accessibility of such solutions across multiple disciplines has been demonstrated. Nevertheless, this research direction remains in its infancy stage, and further explorations are encouraged in other scientific and engineering scenarios.

\section*{Acknowledgment}
Melvin Wong gratefully acknowledges the financial support from Honda Research Institute Europe (HRI-EU). The work is also partially supported by the National Research Foundation (NRF), Singapore, through the AI Singapore Programme under the project titled ``AI-based Urban Cooling Technology Development'' (Award No. AISG3-TC-2024-014-SGKR), as well as by the College of Computing and Data Science (CCDS) at Nanyang Technological University (NTU). The authors would like to thank Mr. Jason Fow Kow, Mr. Lim Rui Yi Ray, Ms. Sun Sitong, and Ms. Swaminathan Navitraa for their assistance in 3D printing the AI-generated designs and supporting video production. Grammarly and ChatGPT were used as writing assistance tools to enhance the readability of the manuscript.

\bibliographystyle{IEEEtran}
\bibliography{References}

\end{document}